\documentclass{article}
\usepackage{arxiv}

\usepackage[utf8]{inputenc} 
\usepackage[T1]{fontenc}    
\usepackage{url}            
\usepackage{booktabs}       
\usepackage{amsmath,amssymb,amsfonts}       
\usepackage{nicefrac}       
\usepackage{microtype}      
\usepackage{graphicx}
\usepackage{multirow,placeins,appendix}
\usepackage[caption=false]{subfig}
\usepackage[numbers,sort&compress]{natbib}
\usepackage[hidelinks]{hyperref}
\usepackage{doi}

\ifdefined\pdfminorversion
\fi

\title{RumorGuard: An Invariant Rationale Guided Graph Contrastive Learning Approach for Robust Rumor Detection}
\author{Yongchao Xia \qquad Caiyan Jia\thanks{Corresponding author.}\\
  School of Computer and Information Technology\\
  Beijing Key Laboratory of Traffic Data Analysis and Mining\\
  Beijing Jiaotong University, Beijing 100044, China\\
  \texttt{21112018@bjtu.edu.cn} \quad \texttt{cyjia@bjtu.edu.cn}}
\date{}
\renewcommand{\shorttitle}{RumorGuard for Robust Rumor Detection}
\hypersetup{
  pdftitle={RumorGuard: An Invariant Rationale Guided Graph Contrastive Learning Approach for Robust Rumor Detection},
  pdfauthor={Yongchao Xia, Caiyan Jia},
  pdfsubject={Rumor detection on social media},
  pdfkeywords={rumor detection, graph contrastive learning, invariant rationale, graph neural networks, interpretability},
}

\begin{document}
\maketitle

\begin{abstract}
Graph contrastive learning (GCL) has achieved strong performance in rumor detection by leveraging graph augmentation. Existing view-generation strategies generally fall into two categories: random augmentation, which risks distorting critical semantic information, and knowledge-guided augmentation, which may introduce inductive biases that degrade model performance. To address these limitations, we propose RumorGuard, a rumor detection framework grounded in the principles of rationale sufficiency and independence. Specifically, RumorGuard employs a rationale-invariant generator to identify salient nodes within rumor propagation graphs and construct rationale-aware augmented views for contrastive learning. This design effectively produces high-quality positive and negative pairs, facilitating more discriminative representation learning. By focusing on informative nodes while filtering out structural and semantic noise, RumorGuard improves detection accuracy while enhancing interpretability. Extensive experiments on five public benchmark datasets demonstrate that RumorGuard consistently outperforms state-of-the-art baselines. Additionally, early detection experiments demonstrate RumorGuard's effectiveness as comments accumulate, while a qualitative case study illustrates how the learned node scores highlight informative replies.
\end{abstract}

\keywords{rumor detection \and graph contrastive learning \and invariant rationale \and graph neural networks \and interpretability}

\section{Introduction}
    In the digital era, the proliferation of social media has facilitated the rapid dissemination of disinformation, degrading online environments and user experiences while undermining trust in social networks. Misinformation and malicious rumors can mislead the public, disrupt individuals' lives, erode social cohesion, and even threaten financial markets and political stability. Consequently, developing effective rumor detection methods is vital for safeguarding public discourse, maintaining social stability, and ensuring the reliability of digital information~\cite{sun2022rumor}.

    Traditional rumor detection approaches primarily rely on static features, such as textual content~\cite{castillo2011information,yang2015emerging}, temporal patterns~\cite{kwon2013prominent,wu2015false}, and user profiles~\cite{yang2012automatic}. However, these features often fail to capture rich semantic nuances and complex structural cues. Subsequent studies have shown that modeling dynamic propagation patterns yields superior performance~\cite{chang2016extreme,kwon2017rumor}. More recently, deep learning techniques, particularly Graph Neural Networks (GNNs)~\cite{bian2020rumor,min2022divide}, have significantly advanced rumor detection by explicitly modeling information diffusion~\cite{ma2016detecting,ma2018rumor}. These advances underscore the importance of integrating propagation modeling with deep representation learning.
    
    \begin{figure}[t]
        \centerline{\includegraphics[width=0.6\textwidth]{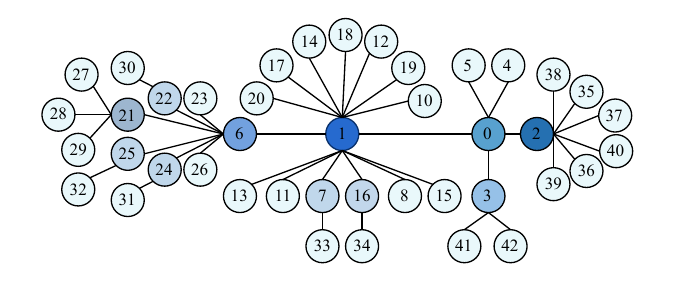}}
        \caption{Visualization of a news propagation graph from Politifact comprising 43 nodes. Node 0 denotes the source post, the remaining nodes represent comments, and edges indicate reply relationships. Node colors reflect varying node degrees.}
        \label{fig1}
    \end{figure}

    Although GNNs perform well on propagation structure learning, they typically assume clean graph inputs and do not explicitly account for irrelevant or noisy comments arising during diffusion, which limits their generalization~\cite{wei2021towards}. To address this challenge, Graph Contrastive Learning (GCL) offers a promising solution. By capturing the intrinsic structural and semantic properties of graphs, GCL reduces sensitivity to noise and produces more robust, transferable representations. Specifically, GCL enhances structural comprehension by contrasting positive and negative view pairs, enabling the model to extract informative patterns while suppressing noisy perturbations~\cite{you2020graph,hassani2020contrastive}.

    In general, GCL-based methods consist of two core components: (1) graph augmentation, which perturbs the topology or node attributes of an anchor graph to generate augmented views; and (2) contrastive learning, which maximizes representation agreement between augmented views of the same anchor graph while separating views from different graphs. Clearly, graph augmentation plays a decisive role in preserving and highlighting discriminative graph characteristics. Existing GCL-based rumor detection methods generally adopt either random augmentation~\cite{bian2020rumor,sun2022rumor} or knowledge-guided augmentation~\cite{cui2024propagation}. However, both strategies exhibit inherent limitations:
    \begin{itemize}
    \item Random augmentation can severely compromise semantic integrity. Unlike images, which often preserve semantics under standard transformations, graphs are highly sensitive to structural and attribute perturbations—even minor alterations can distort their underlying meaning~\cite{yin2022autogcl,cubuk2019autoaugment}. On social media platforms, nodes associated with trending discussions frequently serve as hubs of intensive commentary and interaction (e.g., node 3 in \figurename~\ref{fig1}). Such pivotal nodes and their connections reflect user engagement intensity and delineate emotional transmission pathways. Randomly removing these nodes can disrupt information flow, impeding the reconstruction of communication patterns and emotional dynamics, and ultimately degrading detection accuracy.
    \item Knowledge-guided augmentation preserves domain priors~\cite{cui2024propagation}, yet it risks introducing inductive biases~\cite{li2022let}. For instance, Cui {\it et al.}~\cite{cui2024propagation} propose an adaptive augmentation strategy based on node centrality to improve rumor detection. However, relying heavily on heuristics such as degree centrality may cause low-degree nodes (e.g., node 6 in \figurename~\ref{fig1}) to be mistakenly discarded as unimportant, even when they act as critical bridges in the propagation tree. Over-reliance on predefined priors can therefore bias view generation and impair model flexibility and accuracy.
    \end{itemize}

    \textbf{Graph Contrastive Learning.} Given augmented views of an anchor graph $g$, contrastive learning formulates an instance-discrimination objective as follows:
    \begin{equation}
        \min _{f} \mathcal{L}_{\mathrm{CL}}=\mathbb{E}_{g \in \mathcal{G}} \mathbb{E}_{A_{1}, A_{2} \sim \mathcal{A}} l_{c}\left(f\left(A_{1}(g)\right), f\left(A_{2}(g)\right)\right),
    \label{eq_1}
    \end{equation}
    where $f(\cdot)$ is a GNN encoder that generates graph representations, and $A_{1}, A_{2}$ are two augmentation functions randomly selected from the augmentation set $\mathcal{A}$. The contrastive loss function $l_{c}(\cdot, \cdot)$ can take various forms, such as InfoNCE~\cite{oord2018representation} and NT-Xent~\cite{chen2020simple}. GCL aims to ensure that different augmented views of the same graph yield consistent representations in the feature space, meaning that $f(A(g))$ preserves the invariant features of graph $g$ regardless of the augmentation methods used.
    
    \textbf{Invariant Rationale.} To achieve interpretability in supervised learning~\cite{ying2019gnnexplainer,chen2018learning}, Invariant Rationale (IR)~\cite{wang2022reinforced,li2022invariant} aims to identify a compact subset of input features that serves as the explanatory basis for model predictions. IR is generally governed by two principles~\cite{wu2022handling,li2022let,wu2022discovering}:
    \begin{itemize}
        \item Sufficiency Principle: $p_{\boldsymbol{Y}}(\cdot \mid \boldsymbol{R})=p_{\boldsymbol{Y}}(\cdot \mid \boldsymbol{X})$.
        \item Independence Principle: $\boldsymbol{Y} \bot \boldsymbol{C} \mid \boldsymbol{R}$.
    \end{itemize}
    where $\boldsymbol{X}$, $\boldsymbol{R}$, $\boldsymbol{C}$, and $\boldsymbol{Y}$ denote the original input, the rationale, the rationale's complement, and the label variable, respectively. The Sufficiency Principle states that the rationale $\boldsymbol{R}$ retains sufficient information from $\boldsymbol{X}$ to predict $\boldsymbol{Y}$, whereas the Independence Principle ensures that $\boldsymbol{Y}$ is conditionally independent of $\boldsymbol{C}$ given $\boldsymbol{R}$.

    Motivated by GCL and IR, we posit that effective graph augmentation views should be instance-relevant. To this end, we propose RumorGuard, a graph contrastive learning framework guided by invariant rationales for rumor detection. RumorGuard comprises two key modules: a rationale generator and a rationale-aware contrastive learner. Specifically, the rationale generator assigns an importance score to each node in a rumor propagation graph to extract a discriminative rationale subgraph while filtering out noise. For each anchor graph, the contrastive learner constructs two rationale views as positive pairs and a complement view (along with rationale views from other graphs) as negative samples for instance discrimination. By integrating these two modules, RumorGuard learns discriminative and interpretable representations of rumor propagation graphs. Extensive experiments across multiple benchmark datasets demonstrate that RumorGuard consistently outperforms existing GCL-based rumor detection methods.

\section{Related Work}
    In this section, we review existing literature on social media rumor detection and recent advances in graph contrastive learning.

\subsection{Social Media Rumor Detection}
    Social media rumor detection methods can be broadly categorized into four main paradigms: content-based, user-based, structure-based, and multimodal approaches.

    Content-based methods primarily leverage textual content to verify event authenticity. Early studies rely on handcrafted features to train supervised classifiers, such as Decision Trees~\cite{castillo2011information} and Support Vector Machines~\cite{yang2012automatic}, incorporating cues like user comments~\cite{giudice2010crowdsourcing}, emotional stances~\cite{liu2015real}, and topic distributions~\cite{castillo2011information}. However, feature engineering is labor-intensive and struggles to capture deep semantic and propagation patterns. To overcome these limitations, deep learning models have been adopted to learn complex representations automatically. For instance, Recurrent Neural Networks (RNNs)~\cite{ma2016detecting} model temporal sequences along propagation paths but overlook structural dispersion, while tree-structured Recursive Neural Networks (RvNNs)~\cite{ma2018rumor} capture hierarchical reply relationships. Convolutional Neural Networks (CNNs)~\cite{yu2017convolutional,yu2019attention} extract local neighborhood features to reflect local dispersion, yet remain limited in modeling global structural dependencies~\cite{bruna2013spectral}.

    User-based methods focus on modeling user characteristics and behavioral patterns. Yang {\it et al.}~\cite{yang2012automatic} analyze profile attributes such as account verification, gender, location, and follower count. Wang {\it et al.}~\cite{wang2019spotting} observe notable behavioral differences between users who spread fake news and those who share real news. Jiang {\it et al.}~\cite{jiang2018identifying} propose a session-based graph embedding approach to identify user traits across shared accounts. Although user features can enhance detection accuracy, these methods rely on user metadata that is often incomplete or unavailable due to privacy restrictions.

    Structure-based approaches exploit propagation topologies to characterize how events diffuse across social media. Bian {\it et al.}~\cite{bian2020rumor} capture both causal propagation along reply chains via top-down aggregation and community-level dispersion via bottom-up aggregation. Multi-relational graph embedding methods~\cite{wang2019spotting} further model interactions among heterogeneous entities associated with source posts. Owing to their strong empirical performance, structure-based methods have become the mainstream paradigm for rumor detection.

    Multimodal methods integrate semantic information from both text and visual content to improve detection. Wang {\it et al.}~\cite{wang2018eann} propose a framework that pairs a multimodal feature extractor with an event discriminator to remove event-specific artifacts and learn event-invariant representations. Zhang {\it et al.}~\cite{zhang2020multimodal} introduce a multimodal decoupled domain adaptation framework that disentangles posts into event content and shared rumor style across events. Combining textual and visual modalities enriches feature representations and strengthens discriminative performance.

\subsection{Graph Contrastive Learning}
    In recent years, graph contrastive learning (GCL) has attracted substantial attention within self-supervised learning, with representative GCL methods~\cite{hassani2020contrastive,you2020graph} matching or even surpassing traditional supervised approaches. By minimizing a contrastive loss~\cite{hassani2020contrastive}, augmented views of the same anchor graph (positive pairs) are pulled closer in the representation space, while views from different anchor graphs (negative pairs) are pushed apart, enabling the encoder to learn invariant features for downstream tasks. Most existing GCL methods construct augmented views by applying stochastic perturbations to graph topology or node attributes. For example, Hassani {\it et al.}~\cite{hassani2020contrastive} contrast representations across multiple topological views, while Zhu {\it et al.}~\cite{zhu2020deep} introduce a hybrid scheme that perturbs both topology and node attributes. Despite their effectiveness, random perturbations risk distorting critical semantics and discarding informative structures.

    As mentioned above, data augmentation is crucial to GCL, and existing studies~\cite{yin2022autogcl} show that effective augmentation strategies are highly task-dependent. In computer vision, Cubuk {\it et al.}~\cite{cubuk2019autoaugment} pioneer automated image augmentation via reinforcement learning. In the graph domain, Yin {\it et al.}~\cite{yin2022autogcl} learn task-adaptive graph augmentations guided by the InfoMax~\cite{velickovic2019deep} and InfoMin~\cite{tian2020makes} principles. Specifically for rumor detection, knowledge-guided augmentation methods have been proposed to mitigate the limitations of random perturbations. For example, Cui {\it et al.}~\cite{cui2024propagation} construct salient subgraphs by combining root-node importance with node centrality measures. Nevertheless, over-reliance on predefined prior knowledge can introduce inductive biases that impair model adaptability and accuracy. Consequently, automatically extracting invariant rationales directly from rumor propagation graphs offers a principled way to overcome the limitations of existing GCL augmentation strategies.

\begin{figure}[!t]
    \centering
    \includegraphics[width=1.0\linewidth]{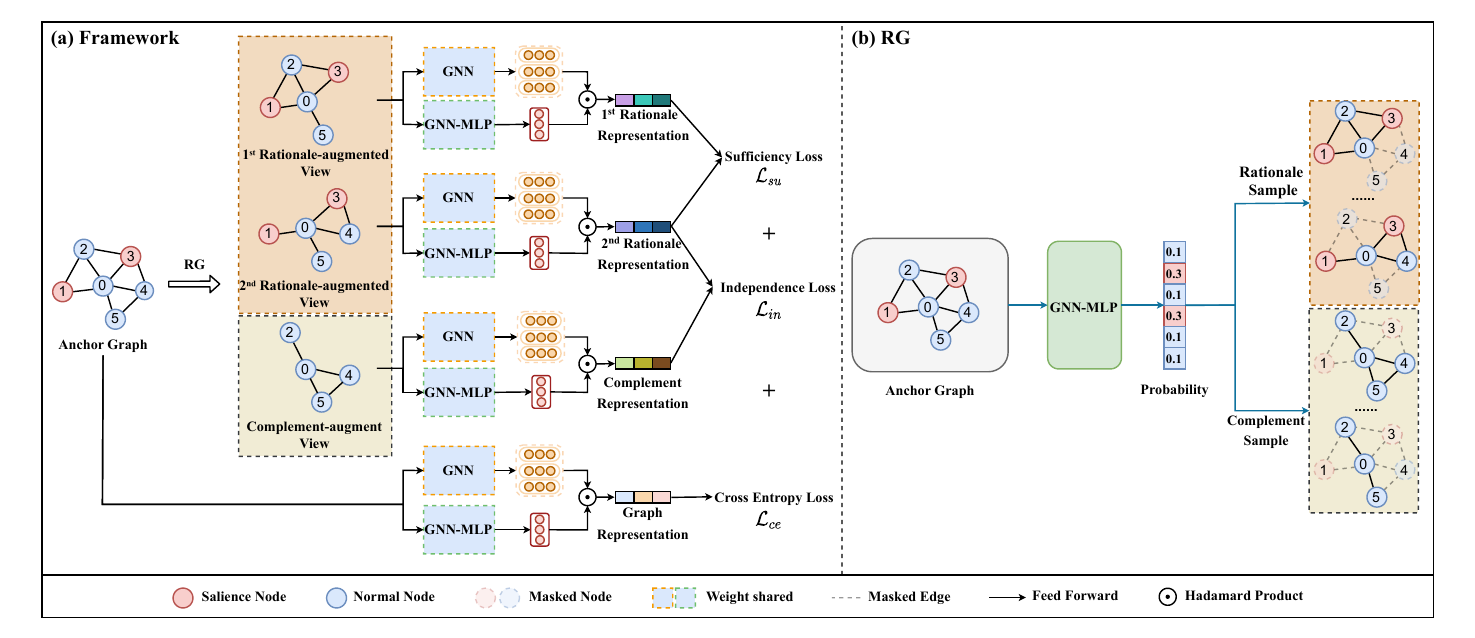}
    \caption{(a) Overview of the RumorGuard framework. The rationale generator identifies a subset of nodes most relevant for event classification. Red nodes retain key semantic information and are sampled to construct rationale-augmented views, whereas blue nodes represent semantically independent nodes sampled to form the complement view. Guided by the sufficiency and independence principles, the contrastive learner performs instance discrimination for each anchor graph using the two rationale views as positive pairs, and both the complement view and rationale views from other graphs as negative samples. Model parameters are jointly optimized by minimizing the sufficiency, independence, and cross-entropy losses. (b) Architecture of the rationale generator, which employs a GNN-MLP encoder to compute an importance score for each node.} \label{fig_method}
\end{figure}

\section{Method}
    In this section, we present RumorGuard, a graph contrastive learning approach guided by invariant rationales. The core idea of RumorGuard is to generate task-relevant views without distorting the salient discriminative features of rumor propagation graphs. As illustrated in \figurename~\ref{fig_method}, we detail the main components of RumorGuard, including a rationale-invariant generator (RG) for rationale-aware graph augmentation and a rationale-aware contrastive learner that learns robust event representations guided by a pretext instance-discrimination task.

\subsection{Problem Definition}
    Let $\mathcal{D} = \{c_{1}, c_{2}, \ldots, c_{n}\}$ denote a labeled event dataset, where $c_i$ is the $i$-th event and $n$ is the total number of labeled events. Each event $c = (y, G)$ consists of a ground-truth label $y \in \{R, N\}$, indicating whether the event is a rumor ($R$) or a non-rumor ($N$), and a propagation graph $G = (\mathcal{V}, \mathcal{E})$, where $\mathcal{V}$ and $\mathcal{E}$ denote the sets of nodes and edges, respectively. In some benchmarks, rumor detection is formulated as a four-class classification task with $y \in \{N, F, T, U\}$, corresponding to Non-rumor, False Rumor, True Rumor, and Unverified Rumor, respectively. Our goal is to train a classifier $f(\cdot)$ that accurately predicts the veracity labels of unlabeled events.

\subsection{Rationale-aware Graph Augmentation Generator}
    Given an anchor graph $G = g$, its rationale $R(G)$ is characterized
    by a probability distribution $P_R(R(G) \mid G = g)$ that reflects
    the likelihood of each node being salient:
    \begin{align}
        P_R(R(G)&=R(g) \mid G=g)\nonumber\\
        &=\prod_{v \in \mathcal{V}_{R}} p(v \mid g)
          \prod_{v \in \mathcal{V}_{C}} \bigl(1 - p(v \mid g)\bigr),
    \label{eq_2}
    \end{align}
    where $\mathcal{V}$ and $\mathcal{V}_R$ are the node set of $g$ and its rationale subset, respectively; \mbox{$\mathcal{V}_C = \mathcal{V} \setminus \mathcal{V}_R$} is the complementary node set; and $p(v \mid g)$ denotes the probability of node $v$ being included in $R(g)$, reflecting the discriminative information carried by $v$. Analogously, we define the distribution of the complement $C(G)$ of rationale $R(G)$ as follows:
    \begin{align}
        P_{C}(C(G)&=C(g) \mid G=g)\nonumber\\
        &=\prod_{v \in \mathcal{V}_{C}} \bigl(1 - p(v \mid g)\bigr)
          \prod_{v \in \mathcal{V}_{R}} p(v \mid g),
    \label{eq_3}
    \end{align}
    where $1 - p(v \mid g)$ measures the likelihood that node $v$ is uninformative for instance discrimination, with a larger value indicating lower discriminative power.
    
    \figurename~\ref{fig_method}(b) illustrates the proposed architecture for generating rationale and complement views. We parameterize this process with a rationale generator $r(\cdot)$ as follows:
    \begin{equation}
        \mathbf{S} = r(g),
    \label{eq_4}
    \end{equation}
    where $r(\cdot)$ is a GNN-MLP encoder that takes an anchor graph $g$ as input and outputs normalized node scores $\mathbf{S} \in \mathbb{R}^{|\mathcal{V}| \times 1}$. The $i$-th entry of $\mathbf{S}$ represents the probability that node $v_i$ belongs to the rationale subgraph.

    Based on the node scores $\mathbf{S}$, we sample a subgraph $R(g)$ from $g$ to construct a rationale-aware view:
    \begin{equation}
        R(g) \sim P_{R}(\cdot \mid G = g)
        \quad \text{s.t.} \quad
        |\mathcal{V}_{R}| = \rho_r |\mathcal{V}|,
    \label{eq_5}
    \end{equation}
    where $\rho_r \in (0, 1)$ is the rationale sampling ratio. Specifically, we sample $\rho_r |\mathcal{V}|$ nodes from $g$ according to $\mathbf{S}$ while retaining all edges among the sampled nodes. Similarly, we sample a subgraph $C(g)$ as a complement view:
    \begin{equation}
        C(g) \sim P_{C}(\cdot \mid G = g)
        \quad \text{s.t.} \quad
        |\mathcal{V}_{C}| = \rho_c |\mathcal{V}|,
    \label{eq_6}
    \end{equation}
    where $\rho_c \in (0, 1)$ is the complement sampling ratio, so $C(g)$ contains $\rho_c |\mathcal{V}|$ nodes from $g$. Note that $C(g)$ is not simply the set difference between $g$ and $R(g)$, but rather a subgraph independently sampled from $P_C(\cdot \mid G = g)$.

\subsection{Rationale-aware Representation Encoder}
    We represent each event's diffusion process as an undirected propagation graph and encode its edge set $\mathcal{E}$ into an adjacency matrix $\mathbf{A} \in \{0, 1\}^{|\mathcal{V}| \times |\mathcal{V}|}$, where $\mathbf{A}_{i, j} = 1$ if nodes $v_i$ and $v_j$ are connected or $i = j$, and $\mathbf{A}_{i, j} = 0$ otherwise. We then feed the graph into a Graph Convolutional Network (GCN)~\cite{kipf2016semi}, whose layer-wise forward propagation is formulated as:
    \begin{equation}
        \mathbf{H}_{l} = \sigma\bigl(\mathbf{A} \mathbf{H}_{l-1}
        \mathbf{W}_{l-1}\bigr),
    \label{gcn_forward}
    \end{equation}
    where $\mathbf{H}_{l} \in \mathbb{R}^{|\mathcal{V}| \times d^{(l)}}$ denotes the hidden node representations at the $l$-th layer, $\mathbf{W}_{l-1} \in \mathbb{R}^{d^{(l-1)} \times d^{(l)}}$ is the trainable weight matrix, and $\sigma(\cdot)$ denotes the ReLU activation function.

    Since source posts contain rich semantic cues, prior studies~\cite{bian2020rumor,he2021rumor} demonstrate that strengthening root-node information for other nodes benefits rumor detection. To this end, we adopt a root-feature augmentation strategy, which concatenates the node features at layer $l$ with the root node representation from layer $l - 1$:
    \begin{equation}
        \tilde{\mathbf{H}}_{l} = \operatorname{concat}\Bigl(
        \mathbf{H}_{l},\, \bigl(\mathbf{H}_{l-1}\bigr)^{\text{root}}
        \Bigr).
    \label{root_aug}
    \end{equation}
    After obtaining the node embeddings $\tilde{\mathbf{H}}_{k}$ at the final ($k$-th) layer, we weight each node representation by its importance score in $\mathbf{S}$ via element-wise multiplication and apply a $\operatorname{Pooling}(\cdot)$ operator to aggregate them into a graph-level representation:
    \begin{equation}
        \mathbf{x} = \operatorname{Pooling}\bigl(\tilde{\mathbf{H}}_{k}
        \odot \mathbf{S}\bigr).
    \label{pool}
    \end{equation}
    
    In this manner, we obtain the graph representations $\mathbf{x}_o$, $\mathbf{x}_r$, and $\mathbf{x}_c$ for the anchor graph, its rationale view, and its complement view, respectively. Furthermore, we apply a projection head $h(\cdot)$ to $\mathbf{x}_r$ and $\mathbf{x}_c$ to map them into a shared latent space for contrastive learning:
    \begin{align}
        \mathbf{r} &= h(\mathbf{x}_{r})
        = L_{2}\bigl(\operatorname{MLP}(\mathbf{x}_{r})\bigr),
        \label{eq:rationale-projection}\\
        \mathbf{c} &= h(\mathbf{x}_{c})
        = L_{2}\bigl(\operatorname{MLP}(\mathbf{x}_{c})\bigr),
        \label{unsup_res}
    \end{align}
    where $h(\cdot)$ is implemented as a multi-layer perceptron (MLP) with $L_2$-normalized outputs.

\subsection{Rationale-aware Contrastive Learner}
    Given an anchor graph $g$, we sample a pair of rationale-aware views $\bigl(R_{1}(g), R_{2}(g)\bigr)$ and obtain their projected representations $\mathbf{r}_{1}^{+}$ and $\mathbf{r}_{2}^{+}$ via Eq.~\eqref{eq:rationale-projection}. Treating the rationale views of other anchor graphs in the mini-batch as negative samples $\mathcal{R}_{g}^{-}$, we operationalize the sufficiency principle by minimizing the following contrastive loss:
    \begin{equation}
        \mathcal{L}_{\mathrm{su}}(g) = -\log
        \frac{\exp\bigl(\mathbf{r}_{1}^{+\top}\mathbf{r}_{2}^{+} / \tau\bigr)}
        {\sum_{\mathbf{r}^{-} \in \mathcal{R}_{g}^{-}}
        \exp\bigl(\mathbf{r}_{1}^{+\top}\mathbf{r}^{-} / \tau\bigr)},
    \label{inv_loss}
    \end{equation}
    where $\tau$ is a temperature hyperparameter. This objective encourages consistency between rationale views of the same anchor graph while pushing apart those from different graphs, thereby guiding the rationale generator to extract discriminative semantics for instance discrimination.

    Furthermore, we sample a complement view $C(g)$ for each anchor graph as an additional negative sample and obtain its projected representation $\mathbf{c}$ via Eq.~\eqref{unsup_res}. The independence principle is then formalized through the following contrastive loss:
    \begin{equation}
        \mathcal{L}_{\mathrm{in}}(g) = -\log
        \frac{\exp\bigl(\mathbf{r}_{1}^{+\top}\mathbf{r}_{2}^{+} / \tau\bigr)}
        {\exp\bigl(\mathbf{r}_{1}^{+\top}\mathbf{r}_{2}^{+} / \tau\bigr)
        + \sum_{\mathbf{c} \in \mathcal{C}}
        \exp\bigl(\mathbf{r}_{1}^{+\top}\mathbf{c} / \tau\bigr)},
    \label{suf_loss}
    \end{equation}
    where $\mathcal{C}$ denotes the set of all complement representations in the mini-batch. Minimizing $\mathcal{L}_{\mathrm{in}}(g)$ pushes each complement representation $\mathbf{c}$ away from the rationale representation $\mathbf{r}_{1}^{+}$, preserving the stability of the extracted rationale even under variations in the complement view. Incorporating complement views as negative samples thus further enhances the model's robustness.

\subsection{Rumor Classifier}
    Given the anchor graph representation $\mathbf{x}_o$ of an event $g$, we feed $\mathbf{x}_o$ into a fully connected layer followed by a softmax layer to obtain the predicted class probability distribution:
    \begin{equation}
        \hat{\mathbf{y}} = \operatorname{softmax}\bigl(
        \mathbf{W}^{F} \mathbf{x}_{o} + \mathbf{b}^{F}\bigr),
    \label{cls}
    \end{equation}
    where $\hat{\mathbf{y}} \in \mathbb{R}^{C}$ denotes the predicted probability distribution over $C$ classes, and $\mathbf{W}^{F}$ and $\mathbf{b}^{F}$ are the trainable weight matrix and bias vector, respectively. Finally, we minimize the cross-entropy loss between the predictions and the ground-truth labels as follows:
    \begin{equation}
        \mathcal{L}_{\mathrm{ce}}(g) = -\sum_{j=1}^{C}
        y_{j} \log \hat{y}_{j},
    \label{eq_19}
    \end{equation}
    where $y_{j} \in \{0, 1\}$ is the $j$-th entry of the one-hot ground-truth label vector of event $g$, and $\hat{y}_{j}$ is the corresponding predicted probability.

\subsection{Model Training}
    We train RumorGuard end-to-end by jointly minimizing the cross-entropy classification loss $\mathcal{L}_{\mathrm{ce}}(g)$ and the rationale-aware contrastive losses $\mathcal{L}_{\mathrm{in}}(g)$ and $\mathcal{L}_{\mathrm{su}}(g)$:
    \begin{equation}
        \min_{\Theta} \mathcal{L} = \mathbb{E}_{g \in \mathcal{G}}
        \Bigl[\mathcal{L}_{\mathrm{ce}}(g)
        + \mu \, \mathcal{L}_{\mathrm{in}}(g)
        + \lambda \, \mathcal{L}_{\mathrm{su}}(g)\Bigr],
    \label{joint objective}
    \end{equation}
    where $\Theta$ denotes all trainable parameters of the model (including the rationale generator, GCN encoder, projection head, and classifier), and $\mu$ and $\lambda$ are trade-off hyperparameters controlling the contributions of $\mathcal{L}_{\mathrm{in}}(g)$ and $\mathcal{L}_{\mathrm{su}}(g)$, respectively.

\section{Experiments}
    In this section, we first evaluate the empirical performance of RumorGuard against representative baseline models. Second, we conduct an ablation study to verify the contribution of each component. Third, we evaluate RumorGuard on early rumor detection tasks to assess its ability to extract discriminative rationales under limited observations. Finally, we present a case study to illustrate the interpretability of the proposed model.
    
\subsection{Datasets}
    We evaluate RumorGuard on five real-world benchmark datasets: Weibo~\cite{ma2016detecting}, Weibo21~\cite{nan2021mdfend}, Twitter15~\cite{ma2017detect}, Twitter16~\cite{ma2017detect}, and Politifact~\cite{shu2020fakenewsnet}. Weibo, Weibo21, and Politifact contain binary labels: False Rumor (F) and True Rumor (T). In contrast, Twitter15 and Twitter16 contain four classes: Non-rumor (N), False Rumor (F), True Rumor (T), and Unverified Rumor (U). Detailed dataset statistics are provided in Appendix~\ref{sec:dataset Statistics}.

\subsection{Baselines}
    We compare the proposed method with the following state-of-the-art baselines.
    \begin{itemize}
        \item \textbf{RvNN}~\cite{ma2018rumor}: A tree-structured
        recursive neural network equipped with GRUs to learn rumor
        representations along propagation trees.
        \item \textbf{BiGCN}~\cite{bian2020rumor}: A GCN-based model
        that jointly captures top-down propagation and bottom-up
        dispersion to model the global structure of rumor trees.
        \item \textbf{GACL}~\cite{sun2022rumor}: A GCN-based model that
        combines adversarial and contrastive learning to resist
        content-based noise.
        \item \textbf{EBGCN}~\cite{wei2021towards}: A probabilistic
        model that captures propagation uncertainty by encoding
        propagation trees with edge-enhanced Bayesian networks.
        \item \textbf{RDEA}~\cite{he2021rumor}: A GCL-based model that
        employs three graph augmentation strategies to enhance event
        representations.
        \item \textbf{RDSPU}~\cite{zheng2023rumor}: A model that infers
        user propensity from historical posts and incorporates social
        interactions.
        \item \textbf{PSGT}~\cite{zhu2024propagation}: A
        Transformer-based model guided by the information bottleneck
        principle to filter out irrelevant information.
        \item \textbf{RAGCL}~\cite{cui2024propagation}: A GCL-based
        model that incorporates prior knowledge via node centrality
        measures to evaluate node importance.
    \end{itemize}
    Detailed descriptions of the experimental setup and evaluation metrics are provided in Appendix~\ref{sec:Experimental Setup}.

\subsection{Experimental Results}
\begin{table}[t]
    \centering
    \caption{Rumor detection results on the Twitter15 dataset.
    The best results are in \textbf{bold} and the second-best are
    \underline{underlined}.}
    \label{table_twitter15_result}
    \begin{tabular}{lccccc}
        \toprule
        \multirow{2}{*}[-2.5pt]{Method} & \multirow{2}{*}[-2.5pt]{Acc}
        & N & F & T & U \\
        \cmidrule(lr){3-6}
        & & $F_1$ & $F_1$ & $F_1$ & $F_1$ \\
        \midrule
        RvNN       & 0.7922 & 0.8097 & 0.7902 & 0.8258 & 0.7422 \\
        BiGCN      & 0.8372 & 0.8174 & 0.8369 & 0.8938 & 0.8008 \\
        GACL       & 0.8689 & 0.8967 & 0.8584 & 0.8898 & 0.8296 \\
        RDEA       & 0.8750 & 0.8909 & \underline{0.8773} & \underline{0.9047} & 0.8258 \\
        EBGCN      & 0.8054 & 0.7627 & 0.8164 & 0.8649 & 0.7772 \\
        RDSPU      & 0.8682 & \textbf{0.9106} & 0.8532 & 0.8815 & 0.8276 \\
        PSGT       & 0.8378 & 0.8766 & 0.8112 & 0.8631 & 0.8008 \\
        RAGCL      & \underline{0.8777} & 0.8967 & 0.8749 & 0.8988 & \underline{0.8389} \\
        \midrule
        \textbf{RumorGuard} & \textbf{0.8831} & \underline{0.9079} & \textbf{0.8817} & \textbf{0.9048} & \textbf{0.8390} \\
        \bottomrule
    \end{tabular}
\end{table}

\begin{table}[t]
    \centering
    \caption{Rumor detection results on the Twitter16 dataset.
    The best results are in \textbf{bold} and the second-best are
    \underline{underlined}.}
    \label{table_twitter16_result}
    \begin{tabular}{lccccc}
        \toprule
        \multirow{2}{*}[-2.5pt]{Method} & \multirow{2}{*}[-2.5pt]{Acc}
        & N & F & T & U \\
        \cmidrule(lr){3-6}
        & & $F_1$ & $F_1$ & $F_1$ & $F_1$ \\
        \midrule
        RvNN       & 0.8037 & 0.7608 & 0.7635 & 0.8781 & 0.8170 \\
        BiGCN      & 0.8319 & 0.7566 & 0.8133 & 0.9087 & 0.8505 \\
        GACL       & 0.8614 & \underline{0.8643} & 0.8055 & 0.9195 & 0.8534 \\
        RDEA       & \underline{0.8736} & 0.8540 & 0.8310 & \textbf{0.9293} & \underline{0.8777} \\
        EBGCN      & 0.8331 & 0.7314 & \textbf{0.8318} & 0.9097 & 0.8574 \\
        RDSPU      & 0.8527 & 0.8613 & 0.8188 & 0.8834 & 0.8462 \\
        PSGT       & 0.8159 & 0.8291 & 0.7415 & 0.8876 & 0.8019 \\
        RAGCL      & 0.8687 & 0.8575 & 0.8184 & 0.9258 & 0.8737 \\
        \midrule
        \textbf{RumorGuard} & \textbf{0.8822} & \textbf{0.8765} & \underline{0.8313} & \underline{0.9286} & \textbf{0.8945} \\
        \bottomrule
    \end{tabular}
\end{table}
\begin{table}[t]
    \centering
    \small
    \setlength{\tabcolsep}{4.5pt}
    \caption{Rumor detection results on the Weibo dataset.
    The best results are in \textbf{bold} and the second-best are
    \underline{underlined}.}
    \label{table_weibo_result}
    \begin{tabular}{lccccccc}
        \toprule
        \multirow{2}{*}[-2.5pt]{Method} & \multirow{2}{*}[-2.5pt]{Acc}
        & \multicolumn{3}{c}{F} & \multicolumn{3}{c}{T} \\
        \cmidrule(lr){3-5} \cmidrule(lr){6-8}
        & & Prec. & Rec. & $F_1$ & Prec. & Rec. & $F_1$ \\
        \midrule
        RvNN  & 0.9091 & 0.9267 & 0.8872 & 0.9063 & 0.8937 & 0.9307 & 0.9117 \\
        BiGCN & 0.8958 & 0.8941 & 0.8967 & 0.8951 & 0.8985 & 0.8949 & 0.8964 \\
        GACL  & 0.9421 & 0.9365 & 0.9477 & 0.9420 & 0.9480 & 0.9366 & 0.9422 \\
        RDEA  & 0.9329 & 0.9313 & 0.9343 & 0.9324 & 0.9359 & 0.9315 & 0.9333 \\
        EBGCN & 0.8531 & 0.8924 & 0.8011 & 0.8438 & 0.8227 & 0.9043 & 0.8613 \\
        PSGT  & 0.9353 & 0.9305 & 0.9404 & 0.9352 & 0.9410 & 0.9302 & 0.9354 \\
        RAGCL & \underline{0.9438} & \underline{0.9388} & \underline{0.9490} & \underline{0.9437} & \underline{0.9496} & \underline{0.9388} & \underline{0.9440} \\
        \midrule
        \textbf{RumorGuard} & \textbf{0.9558} & \textbf{0.9463} & \textbf{0.9659} & \textbf{0.9559} & \textbf{0.9658} & \textbf{0.9460} & \textbf{0.9557} \\
        \bottomrule
    \end{tabular}
\end{table}

\begin{table}[t]
    \centering
    \small
    \setlength{\tabcolsep}{4.5pt}
    \caption{Rumor detection results on the Weibo21 dataset.
    The best results are in \textbf{bold} and the second-best are
    \underline{underlined}.}
    \label{table_weibo21_result}
    \begin{tabular}{lccccccc}
        \toprule
        \multirow{2}{*}[-2.5pt]{Method} & \multirow{2}{*}[-2.5pt]{Acc}
        & \multicolumn{3}{c}{F} & \multicolumn{3}{c}{T} \\
        \cmidrule(lr){3-5} \cmidrule(lr){6-8}
        & & Prec. & Rec. & $F_1$ & Prec. & Rec. & $F_1$ \\
        \midrule
        RvNN  & 0.8316 & 0.8237 & 0.7874 & 0.8047 & 0.8382 & 0.8664 & 0.8518 \\
        BiGCN & 0.8054 & 0.7981 & 0.7483 & 0.7723 & 0.8106 & 0.8505 & 0.8300 \\
        GACL  & \underline{0.8623} & 0.8701 & \underline{0.8098} & \underline{0.8385} & \underline{0.8577} & 0.9038 & \underline{0.8800} \\
        RDEA  & 0.8302 & 0.8164 & 0.7960 & 0.8052 & 0.8425 & 0.8573 & 0.8494 \\
        EBGCN & 0.7400 & 0.6910 & 0.7508 & 0.7175 & 0.7901 & 0.7314 & 0.7581 \\
        PSGT  & 0.8269 & \textbf{0.8867} & 0.7034 & 0.7797 & 0.8013 & \textbf{0.9241} & 0.8566 \\
        RAGCL & 0.8530 & 0.8767 & 0.7762 & 0.8233 & 0.8380 & \underline{0.9137} & 0.8742 \\
        \midrule
        \textbf{RumorGuard} & \textbf{0.8711} & \underline{0.8780} & \textbf{0.8227} & \textbf{0.8492} & \textbf{0.8667} & 0.9092 & \textbf{0.8873} \\
        \bottomrule
    \end{tabular}
\end{table}

\begin{table}[t]
    \centering
    \small
    \setlength{\tabcolsep}{4.5pt}
    \caption{Rumor detection results on the Politifact dataset.
    The best results are in \textbf{bold} and the second-best are
    \underline{underlined}.}
    \label{table_Politifact_result}
    \begin{tabular}{lccccccc}
        \toprule
        \multirow{2}{*}[-2.5pt]{Method} & \multirow{2}{*}[-2.5pt]{Acc}
        & \multicolumn{3}{c}{F} & \multicolumn{3}{c}{T} \\
        \cmidrule(lr){3-5} \cmidrule(lr){6-8}
        & & Prec. & Rec. & $F_1$ & Prec. & Rec. & $F_1$ \\
        \midrule
        BiGCN & 0.8790 & 0.8968 & 0.8603 & 0.8776 & 0.8642 & 0.8978 & 0.8802 \\
        GACL  & 0.8791 & 0.8885 & 0.8675 & 0.8741 & 0.8800 & 0.8911 & 0.8825 \\
        RDEA  & 0.8886 & 0.8890 & \underline{0.8919} & 0.8898 & 0.8909 & 0.8847 & 0.8871 \\
        EBGCN & 0.8727 & 0.8750 & 0.8792 & 0.8752 & 0.8766 & 0.8657 & 0.8688 \\
        PSGT  & 0.8677 & 0.8693 & 0.8665 & 0.8657 & 0.8745 & 0.8685 & 0.8690 \\
        RAGCL & \underline{0.8949} & \underline{0.8982} & 0.8911 & \underline{0.8940} & \underline{0.8939} & \underline{0.8984} & \underline{0.8955} \\
        \midrule
        \textbf{RumorGuard} & \textbf{0.9204} & \textbf{0.9204} & \textbf{0.9252} & \textbf{0.9173} & \textbf{0.9178} & \textbf{0.9236} & \textbf{0.9201} \\
        \bottomrule
    \end{tabular}
\end{table}

    Tables~\ref{table_twitter15_result}, \ref{table_twitter16_result}, \ref{table_weibo_result}, \ref{table_weibo21_result}, and \ref{table_Politifact_result} summarize the performance of all compared methods across the five real-world datasets, where bold and underlined entries denote the best and second-best results, respectively. Overall, RumorGuard achieves the highest accuracy on every dataset and leads on most class-wise metrics, demonstrating its strong effectiveness in rumor detection.

    First, GCL-based GCN methods, including GACL, RDEA, RAGCL, and RumorGuard, generally outperform non-contrastive baselines such as BiGCN and RvNN. This observation confirms that graph contrastive learning enhances model robustness against structural and content noise while capturing richer node interactions.

    Second, although the Transformer-based method PSGT offers interpretability, its accuracy lags behind RumorGuard by 4.53\%, 6.63\%, 4.42\%, and 5.27\% on Twitter15, Twitter16, Weibo21, and Politifact, respectively. A plausible reason is that global self-attention across all nodes may introduce irrelevant information from distant replies as propagation trees grow.
    
    Finally, while RAGCL leverages prior knowledge for graph augmentation and achieves competitive results, it still trails RumorGuard across all five datasets. Unlike RAGCL's reliance on fixed centrality heuristics, RumorGuard dynamically adapts to dataset-specific characteristics by generating task-relevant rationale views, enabling more flexible and effective representation learning.

    In summary, RumorGuard captures discriminative propagation patterns more effectively than existing baselines, yielding superior detection performance. Moreover, the rationale nodes identified by our model enhance interpretability, as illustrated in our case study (see Appendix~\ref{sec:Case Study}).
    
\subsection{Ablation Study}
\begin{table}[t]
        \centering
        \caption{Ablation study results (accuracy) across five datasets.
        The best results are in \textbf{bold}.}
        \label{ablation_study_table}
        \begin{tabular}{lccccc}
            \toprule
            Variant & Twitter15 & Twitter16 & Weibo & Weibo21 & Politifact \\
            \midrule
            w/o Su & 0.8655 & 0.8626 & 0.9505 & 0.8585 & 0.8220 \\
            w/o In & 0.8784 & 0.8749 & 0.9546 & 0.8697 & 0.8220 \\
            \midrule
            \textbf{RumorGuard} & \textbf{0.8831} & \textbf{0.8822} & \textbf{0.9558} & \textbf{0.8711} & \textbf{0.9204} \\
            \bottomrule
        \end{tabular}
    \end{table}

    In this subsection, we evaluate the contribution of each key component in RumorGuard by comparing two ablation variants: (1) \textbf{w/o Su}, which removes the sufficiency contrastive loss $\mathcal{L}_{\mathrm{su}}$, and (2) \textbf{w/o In}, which removes the independence contrastive loss $\mathcal{L}_{\mathrm{in}}$. As shown in Table~\ref{ablation_study_table}, removing either loss consistently degrades performance across all five datasets, confirming that both principles are essential to RumorGuard.

    Specifically, \textbf{w/o Su} causes a larger performance drop than \textbf{w/o In} on Twitter15, Twitter16, Weibo, and Weibo21, reducing accuracy by 1.76\%, 1.96\%, 0.53\%, and 1.26\%, respectively. This highlights that $\mathcal{L}_{\mathrm{su}}$ plays the primary role in guiding the rationale generator to retain discriminative semantic nodes. Meanwhile, $\mathcal{L}_{\mathrm{in}}$ further improves robustness by pushing complement views away from rationale representations. Notably, on the smaller Politifact dataset, removing either loss leads to a substantial 9.84\% drop in accuracy, indicating that both sufficiency and independence constraints work synergistically to prevent overfitting to noisy replies when training data is limited.

\subsection{Early Rumor Detection}
    \begin{figure}[!t]
        \centering
        \subfloat[Twitter15]{\includegraphics[width=0.48\linewidth]{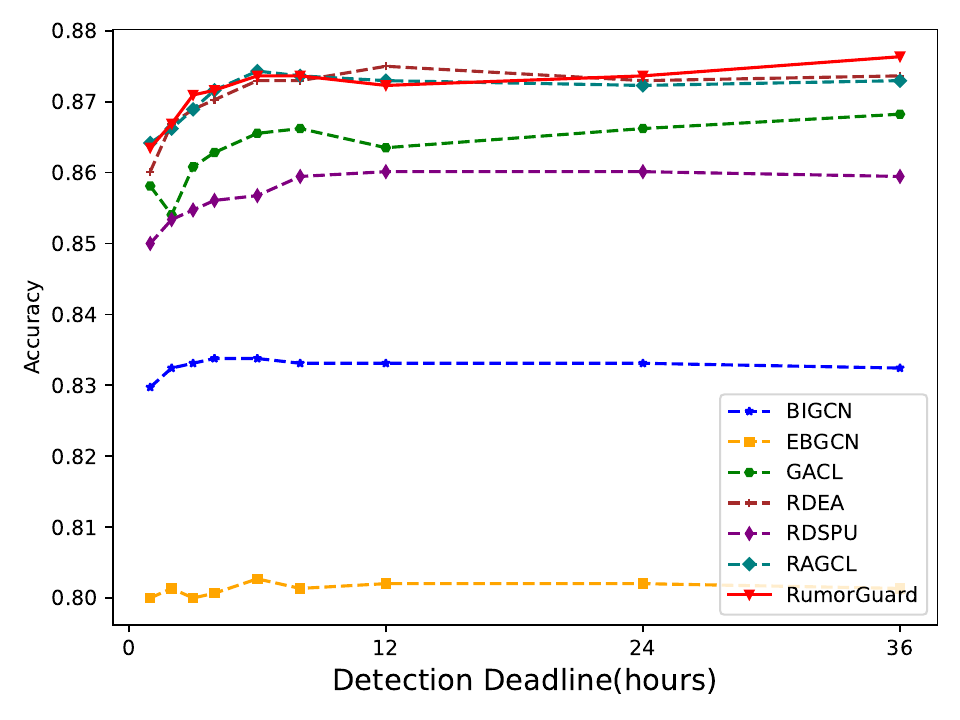}\label{twitter15early}}
        \hfill
        \subfloat[Twitter16]{\includegraphics[width=0.48\linewidth]{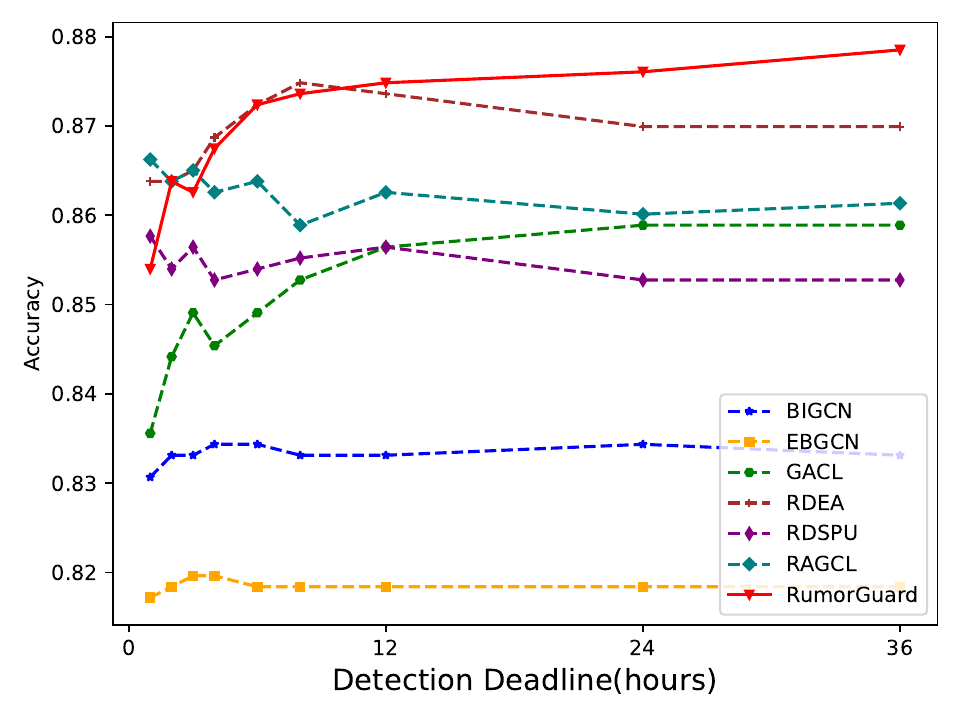}\label{twitter16early}}
        \\[4pt]
        \subfloat[Weibo]{\includegraphics[width=0.48\linewidth]{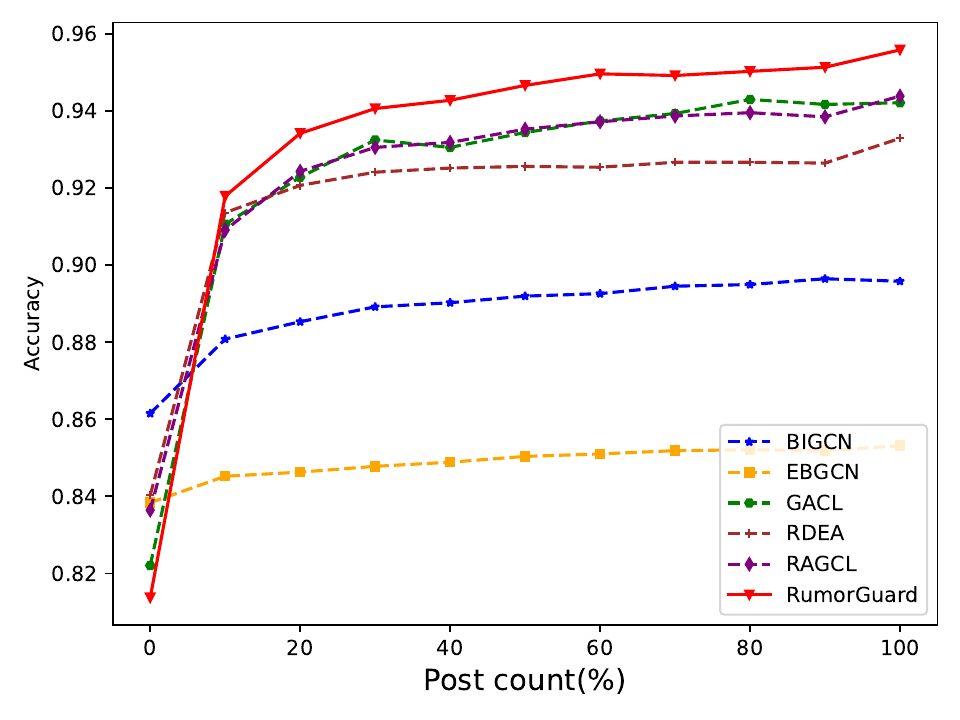}\label{weiboearly}}
        \hfill
        \subfloat[Weibo21]{\includegraphics[width=0.48\linewidth]{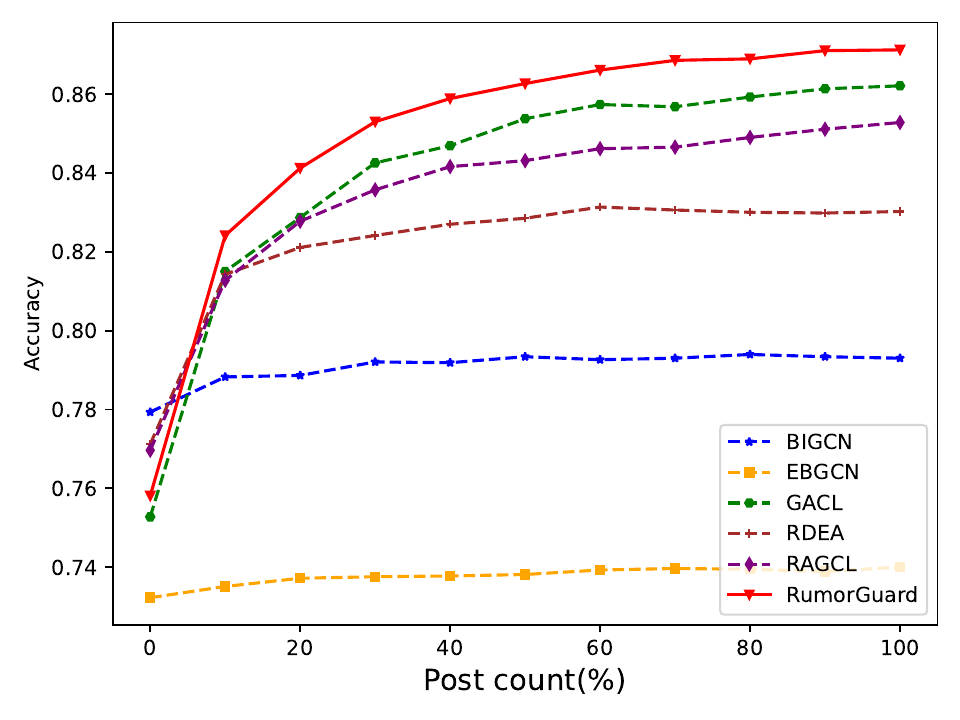}\label{weibo21early}}
        \caption{Early rumor detection results across four datasets.}
        \label{early_detect}
    \end{figure}

    Early rumor detection, which aims to identify rumors during the initial stages of propagation, is another critical criterion for evaluating model effectiveness. To construct the early detection tasks, we set a series of observation deadlines and restrict each model's input to the source post and comments published prior to each deadline. Specifically, we evaluate performance across varying elapsed times since the source post's publication. Note that Politifact is excluded from this evaluation due to the absence of post timestamps.

    As illustrated in \figurename~\ref{early_detect}, RumorGuard consistently outperforms the baseline models across all early detection stages. At time point 0, when only the source post is available, all models exhibit relatively low accuracy due to limited contextual information. As comments accumulate, detection performance improves substantially. However, several baselines, such as RAGCL, RDSPU, and RDEA on Twitter16, experience a slight performance decline after 12 hours, likely because later comments introduce additional structural and semantic noise. In contrast, RumorGuard maintains stable performance as more comments arrive, demonstrating strong robustness against noisy interactions. Overall, results across the four real-world datasets confirm that RumorGuard excels at both general and early rumor detection.

\section{Conclusion}
    In this study, we propose RumorGuard, a rumor detection framework that integrates graph contrastive learning with invariant rationale identification. Specifically, RumorGuard employs a rationale generator to construct task-relevant rationale views that capture critical semantic cues in propagation graphs without relying on predefined prior knowledge. Guided by the sufficiency and independence principles, the proposed method generates augmented views that preserve essential semantics while facilitating effective contrastive learning. Extensive experiments on five benchmark rumor detection datasets demonstrate that our rationale generator enhances both interpretability and generalization, enabling RumorGuard to consistently outperform state-of-the-art baselines.

\section*{Statements and Declarations}

\paragraph*{Funding.}
This work was supported by the National Natural Science Foundation of China (NSFC) under Grant No. 62576026 and the Fundamental Research Funds for the Central Universities (No. 2024XKRC024).

\paragraph*{Competing Interests.}
The authors have no competing interests to declare that are relevant to the content of this article.

\paragraph*{Data Availability.}
All benchmark datasets used in this study are publicly available from their respective original sources, including Weibo~\cite{ma2016detecting}, Weibo21~\cite{nan2021mdfend}, Twitter15 and Twitter16~\cite{ma2017detect}, and the Politifact subset of FakeNewsNet~\cite{shu2020fakenewsnet}. The reconstructed Weibo21 propagation graphs (detailed in Appendix~\ref{sec:dataset Statistics}) are available from the corresponding author upon reasonable request.

\paragraph*{Author Contributions.}
Yongchao Xia conceived the study, designed the methodology, conducted the experiments, and wrote the original draft of the manuscript. Caiyan Jia supervised the research, provided guidance on the methodology and experimental design, and reviewed and edited the manuscript. All authors read and approved the final manuscript.

\bibliographystyle{unsrtnat}
\bibliography{references}

\clearpage
\appendix

\section{Implementation Details}

\subsection{Dataset and Graph Statistics} \label{sec:dataset Statistics}
    Table~\ref{dataset table} summarizes the statistics of the five benchmark datasets used in our experiments. Notably, Weibo21 is a multi-domain Chinese fake news dataset spanning nine categories, originally comprising 9,128 events without propagation structures. To construct propagation graphs for Weibo21, we used the post IDs provided by the original authors and retrieved reply threads via the Sina Weibo API. Owing to platform restrictions and deleted posts, the reconstructed Weibo21 dataset contains 5,266 events with complete propagation structures. For initial node representations, we use 768-dimensional BERT embeddings on Politifact and the top 5,000 TF-IDF word features on the remaining four datasets.
    
    \begin{table}[!htbp]
        \centering
        \caption{Statistics of the five benchmark datasets.}
        \label{dataset table}
        \begin{tabular}{lccccc}
            \toprule
            Statistic & Twitter15 & Twitter16 & Weibo & Weibo21 & Politifact \\
            \midrule
            \# of events            & 1,490   & 818     & 4,664     & 5,266   & 314 \\
            \# of True rumors       & 374     & 205     & 2,351     & 2,942   & 157 \\
            \# of False rumors      & 370     & 205     & 2,313     & 2,324   & 157 \\
            \# of Unverified rumors & 374     & 203     & --        & --      & --  \\
            \# of Non-rumors        & 372     & 205     & --        & --      & --  \\
            \midrule
            \# of posts             & 331,612 & 204,820 & 3,805,656 & 955,312 & 41,054 \\
            \bottomrule
        \end{tabular}
    \end{table}

\subsection{Experimental Setup} \label{sec:Experimental Setup}
    All methods are implemented in PyTorch on a Linux server equipped
    with NVIDIA GeForce RTX 3090 Ti GPUs. Following the official
    implementations, we tune the hyperparameters of each baseline on
    every dataset. For a reliable evaluation, each dataset is randomly
    partitioned into five equal folds for 5-fold cross-validation, using
    identical splits across all compared models to ensure fair
    comparisons. We report Accuracy (Acc.) across all categories,
    together with class-wise $F_1$ scores (as well as Precision and
    Recall on the binary-class datasets). RumorGuard is trained via
    mini-batch gradient descent using the Adam optimizer~\cite{kingma2014adam}
    with a learning rate of 0.0005 and a batch size chosen from
    \{32, 64, 128\}. The hyperparameters $\lambda$, $\mu$, $\rho_{r}$, and
    $\rho_{c}$ are tuned within the range $[0.1, 1.0]$. Early
    stopping~\cite{yao2007early} is applied if the validation loss does
    not decrease for 10 consecutive epochs.

\section{Additional Experimental Results} \label{sec:More Experiment Results}
\subsection{Case Study}\label{sec:Case Study}
    \begin{figure}[!htbp]
        \centering
        \includegraphics[width=0.9\linewidth,height=0.63\textheight,keepaspectratio]{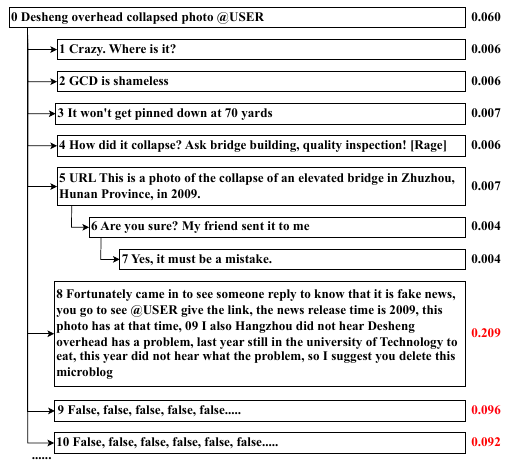}
        \caption{A false rumor sampled from the Weibo21 dataset. Each node represents a post, and a higher score indicates that the post is more informative about the event's authenticity.}
        \label{case study}
    \end{figure}
    In this subsection, we illustrate the interpretability of RumorGuard enabled by the rationale generator through a qualitative case study. As shown in \figurename~\ref{case study}, we examine a false rumor randomly sampled from the Weibo21 dataset, where posts and comments are represented as nodes and reply relationships as edges. Each node is assigned an importance score reflecting its discriminative contribution. Notably, nodes 8, 9, and 10 receive substantially higher scores than the remaining nodes, indicating their decisive role in verifying the event. Furthermore, unlike prior-knowledge-based methods that heavily emphasize the root post~\cite{cui2024propagation}, RumorGuard does not assign the highest score to the source post; instead, it highlights nodes 8, 9, and 10, whose content directly exposes the falsity of the rumor. Overall, the learned node scores demonstrate that RumorGuard effectively pinpoints informative replies, providing strong interpretability alongside accurate detection.

\FloatBarrier
\end{document}